\documentclass[10pt,twocolumn,letterpaper]{article}

\usepackage{3dv}
\usepackage{times}
\usepackage{epsfig}
\usepackage{graphicx}
\usepackage{amsmath}
\usepackage{amssymb}
\usepackage{subfig}
\usepackage{stackengine}

\usepackage{booktabs}
\usepackage{multirow}
\usepackage[table]{xcolor}
\usepackage{setspace}

\makeatletter
\@namedef{ver@everyshi.sty}{}
\makeatother
\usepackage{tikz}
\usetikzlibrary{arrows}
\usetikzlibrary{quotes}
\usetikzlibrary{patterns}
\usetikzlibrary{graphs} 
\usetikzlibrary{shapes.geometric}
\tikzset{font={\fontsize{6pt}{12}\selectfont}}
\tikzstyle{process} = [rectangle, rounded corners, minimum width=0.8cm, minimum height=0.8cm, align=center, draw=black, fill=gray]
\tikzstyle{arrow} = [thick,->,>=stealth]
\tikzset{
  every node/.style={
  , execute at begin node=\setlength{\baselineskip}{1em}
  }
}

\usepackage[pagebackref=true,breaklinks=true,letterpaper=true,colorlinks,bookmarks=false]{hyperref}
\usepackage[numbers]{natbib}

\threedvfinalcopy 

\def\threedvPaperID{86} 

\ifthreedvfinal\fi
\begin{document}

\title{Sparse and Dense Data with CNNs:\\Depth Completion and Semantic Segmentation}


\author{
	Maximilian Jaritz\textsuperscript{1, 2}, Raoul de Charette\textsuperscript{1}, Emilie Wirbel\textsuperscript{2}, Xavier Perrotton\textsuperscript{2}, Fawzi Nashashibi\textsuperscript{1}\\\\
	\and
	\textsuperscript{1}Inria RITS Team\\
	{\tt\small \{maximilian.jaritz, raoul.de-charette, fawzi.nashashibi\}@inria.fr}
	\and
	\textsuperscript{2}Valeo\\
	{\tt\small \{emilie.wirbel, xavier.perrotton\}@valeo.com}
}

\maketitle

\begin{abstract}
Convolutional neural networks are designed for dense data, but vision data is often sparse (stereo depth, point clouds, pen stroke, etc.).
We present a method to handle sparse depth data with optional dense RGB, and accomplish depth completion and semantic segmentation changing only the last layer. 
Our proposal efficiently learns sparse features without the need of an additional validity mask.
We show how to ensure network robustness to varying input sparsities. Our method even works with densities as low as $0.8\%$ (8~layer lidar), and outperforms all published state-of-the-art on the Kitti depth completion benchmark.
\end{abstract}

\section{Introduction}

Most computer vision algorithms rely now on convolutional neural networks (CNNs) which are designed for dense data, rather than sparse data which is rarely considered.
However, vision data can become sparse when reprojected into a different dense coordinate space
(e.g. Lidar data reprojected into camera image plane).
Corrupted data may also be seen as sparse data, either because of noise or invalid measurement (e.g. lack of reflectivity for Lidar, camera saturation).
Finally, some processes are inherently sparse such as stereo disparity that relies on saliency, sparse by nature.

A main objective of processing sparse data is to complete missing information. This is known as data completion (aka inpainting for images) and upscaling is one instance of this problem. 
Classically, this was achieved through sophisticated interpolation of valid data, which failed at completing large holes in data.
Machine learning on the other hand can complete large chunks of missing data from learned appearance priors. 

In robotics, this task has interesting applications, as sensors have different resolutions and field of views, i.e. different densities in a common reference frame. As an example,
the reprojection of a Velodyne 64 layers Lidar only covers 5.9\% pixels of the whole image space in the Kitti dataset and even less density with fewer layers\footnote{3.0\%, 1.6\%, 0.8\% pixels density simulating 32, 16, 8 layers Lidars.}.
Apart from depth completion, the long-term goal of processing sparse data is its fusion with dense data which would benefit all vision tasks in general.




\begin{figure}
	\centering
	\includegraphics[width=0.87\columnwidth]{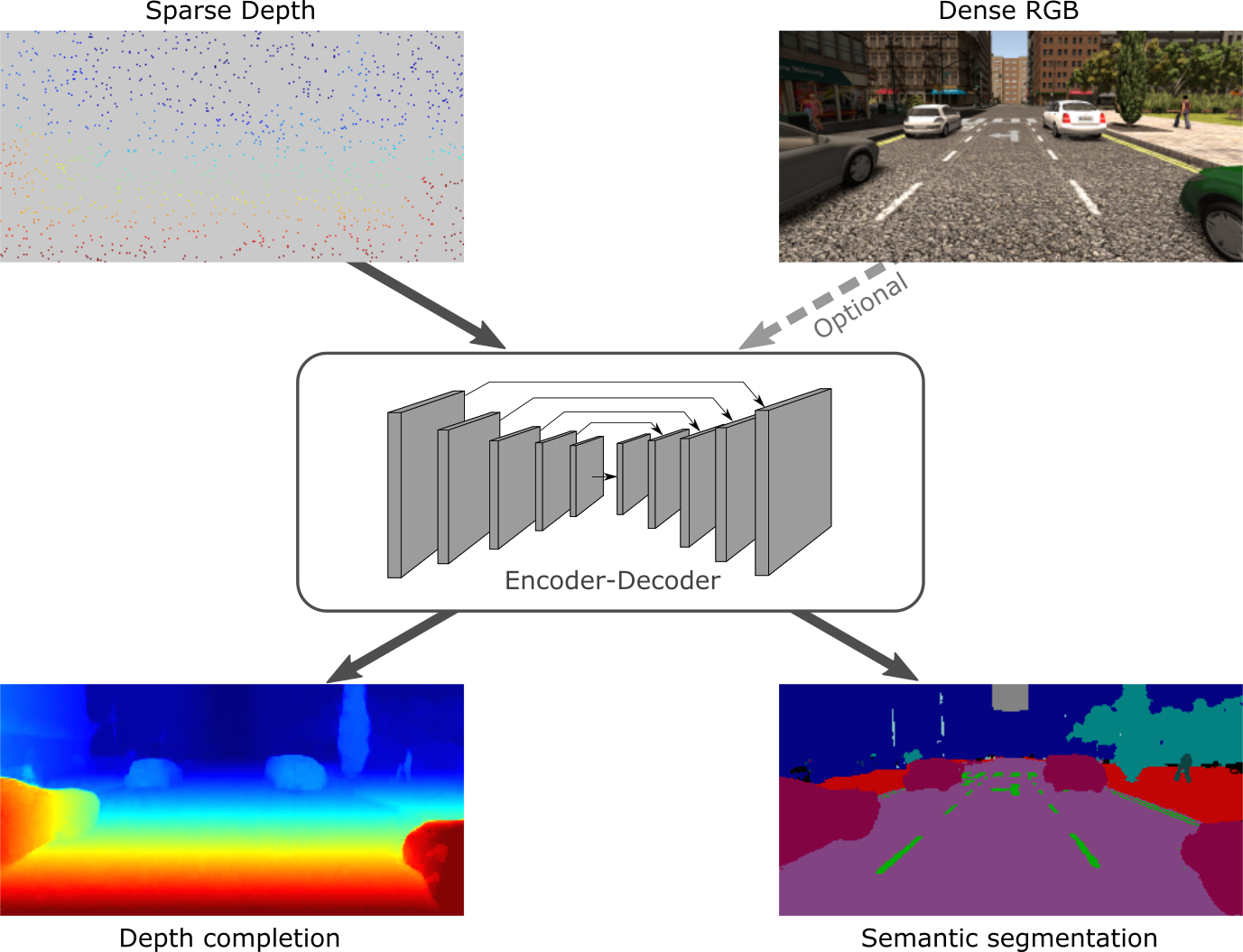}
	\caption{Our method handles sparse depth input data with or without additional dense RGB, to accomplish depth completion or semantic segmentation (with minor adaptation).}
	\label{fig:scheme}
\end{figure}

Naively applying CNN to sparse data does not work as they are sensitive to missing data.
It is commonly claimed that learning all combination of missing data is virtually impossible.
Instead, sparse convolution \cite{uhrig2017sparsity, ren2018SBNet} was proposed recently which allows for a feature representation that is invariant
to missing data using an additional validity mask input.
While this led to an important performance gain, we show that a mask input might be redundant and that using a normal (i.e. dense) CNN architecture
with ad-hoc training process can yield significantly better performance while remaining generic.

Intuitively our results demonstrate that with proper training, CNNs can learn where valid input data are and how to rely on them to build features invariant to the missing data.

\paragraph{Contributions}
We propose a new method to leverage sparse data processing and demonstrate its performance on depth completion and semantic segmentation tasks. Fig.~\ref{fig:scheme} illustrates
our approach: using a common network structure, different kind of data are used to perform inference tasks which require data completion and benefit from data fusion.
Using an encoder-decoder scheme, NASNet \cite{zoph2017learning} with minor changes and a sparse training strategy, we show that our method can efficiently handle sparse inputs of various densities without the need of retraining or any additional mask input. In addition to being lighter and easier to design, our method preserves better sharp edges.
We extend our research to inputs of different densities and prove that dense RGB and sparse depth data can be efficiently fused together in a late fusion manner.
The experiments demonstrate the performance of our approach on depth completion and semantic segmentation for both single sparse input (depth) or sparse + dense inputs (depth + RGB).
Tests are conducted on both synthetic and real data and we carry out an additional ablation study to prove robustness at lower data density. Results on the Kitti Depth Completion benchmark show we outperform state-of-the-art.

\section{Related Work}

\begin{figure}
	\centering
	\subfloat[Structured]{\label{fig:structured-sparsity}\includegraphics[height=0.145\columnwidth,trim={{1.0\columnwidth} 0 {0.4\columnwidth} 0},clip]{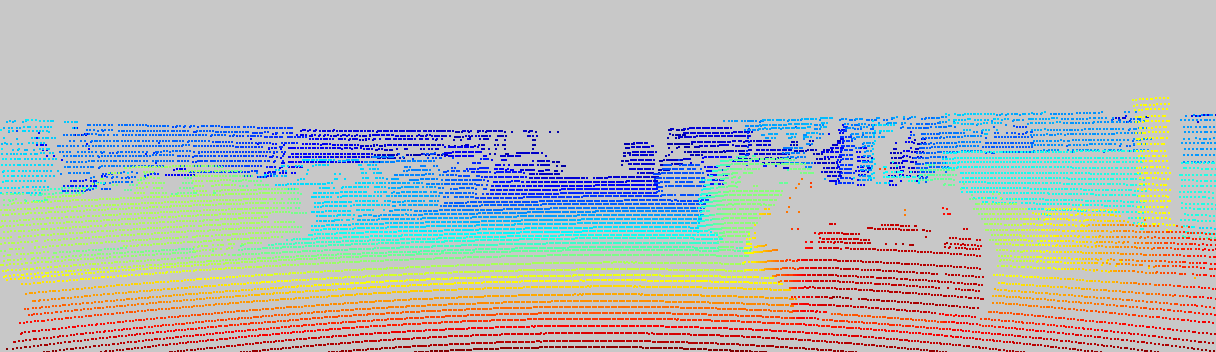}}\hspace{0.01\columnwidth}
	\subfloat[Patches]{\label{fig:almost-dense}\includegraphics[height=0.145\columnwidth]{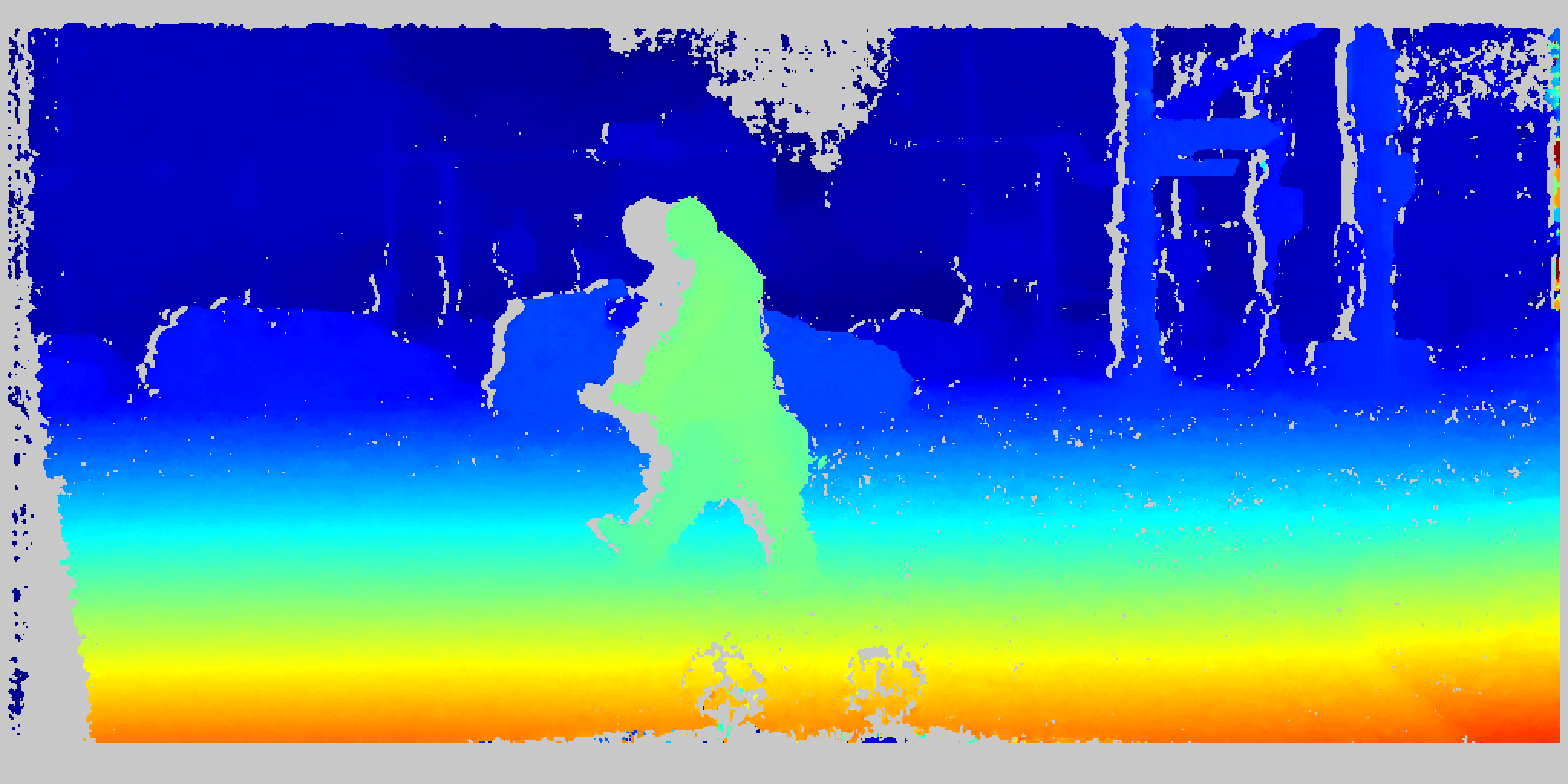}}\hspace{0.01\columnwidth}
	\subfloat[Uniform]{\label{fig:homogeneous}\includegraphics[height=0.145\columnwidth,trim={0 0 0 1.5cm},clip]{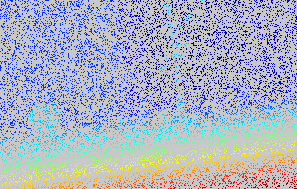}}
	\caption{Different sparsity patterns from sensors such as Lidar \ref{fig:structured-sparsity}, stereo camera \ref{fig:almost-dense} or synthetic data \ref{fig:homogeneous}.}
	\label{fig:sparseDepthExamples}
\end{figure}

We present the state-of-the-art of pixel-wise inference tasks, focusing on Depth Prediction and Semantic Segmentation. 

In the targeted tasks the expected result is a dense output in the image plane and our research relies on past works for dense or sparse processing. Consequently, we first provide the reader with an overview of inference from dense inputs, and then detail inference from sparse inputs only or the fusion of sparse + dense, at the core of this work.

\paragraph{Dense Inputs}

Pixelwise inference tasks such as semantic segmentation seek an output of equal resolution as the input, which led to Fully Convolutional Networks (FCNs). The downsampled CNN features in FCNs are upsampled with bilinear upsampling or transposed convolution \cite{long2015fully}. Skip connections between the equivalent feature maps of the downsampling (encoder) and upsampling (decoder) part of the network allow to keep details \cite{ronneberger2015u, badrinarayanan2017segnet}. 
To preserve spatial context, Pyramid Pooling in PSPNet \cite{zhao2017pyramid} concatenates upsampled local and global multiscale features maps. 
An alternative to downsampling the features is to dilate the convolutional kernels \cite{chen2018deeplab}. This way, skip connections can be omitted as the resolution stays the same.

Multiscale networks are commonly used for semantic segmentation \cite{long2015fully} or depth prediction task \cite{eigen2014depth}. From a single image, depth is inferred with learned priors on object and scenes, for example with VGG \cite{eigen2014depth} or deeper ResNet-50 \cite{laina2016deeper} networks.
Unlike semantic segmentation that requires costly annotation, depth can be easily learned. Either in a supervised manner with sparse Lidar measurements \cite{kuznietsov2017semi} or for example in a self-supervised fashion with stereo by computing the loss between the right camera image and the reprojected version of the left image using depth \cite{godard2017unsupervised}.

\paragraph{Sparse Inputs}
The nature of sparse data varies with the sensor and scenario, as can be seen in fig~\ref{fig:sparseDepthExamples}.
For instance, Lidar data exhibit \textit{structured sparsity} due to the discrete polar scanning behavior (fig.~\ref{fig:structured-sparsity}).
Stereo vision or structured light sensors deliver dense data with \textit{patches} of missing data (fig.~\ref{fig:almost-dense}).
Artificially altered data typically has \textit{uniformly} distributed data (fig.~\ref{fig:homogeneous}).
Classically, sparse 2D inputs are considered for inpainting of \textit{uniform} sparsity using local interpolation \cite{ku2018defense} or guided optimization \cite{silberman2012indoor} for \textit{patches} sparsity.

As CNNs are designed to operate on dense data, a common strategy is to transform sparse data to a 2D or 3D grid with \textit{holes} \cite{uhrig2017sparsity,ren2018SBNet,riegler2017octnet}.
A validity mask can be given as additional input to express valid or missing data \cite{uhrig2017sparsity,ren2018SBNet}. In \cite{uhrig2017sparsity} - which established the groundwork of CNN sparse depth completion - only valid data locations are considered in the convolution and the activation is biased accordingly. As filters do not distinguish between a dense pattern and a pattern with missing values, the network is sparsity independent. Meanwhile, it induces a blurry output due to the extension of the validity domain (see discussion in \ref{sec:validity-mask}). In \cite{ren2018SBNet} the mask is input as block-wise coordinates which leverage the loss of spatial information in \cite{uhrig2017sparsity}, but cannot accommodate to the variety of sparsities.

Other alternatives either act directly on the data manifold \cite{graham2017submanifold} to avoid the extension of the valid domain or apply order-invariant operations \cite{qi2017pointnet}.

\paragraph{Sparse + Dense Inputs}
The problem of inputs with different densities/sparsities is yet little addressed.
For the inference tasks, the sparse depth is either used to guide the RGB inference \cite{zhang2018deep} or just combined \cite{ma2018sparse}.
In \cite{zhang2018deep}, surface normals and occlusion boundaries are inferred from RGB only which provides a coarse geometric representation of the scene which is then completed via a global optimization guided by the sparse depth input. 
Alternatively, \cite{ma2018sparse} concatenates RGB and sparse depth to a 4-channel input to an encoder-decoder network.

\section{Method}
\label{sec:method}

Our aim is to propose a method that efficiently handles sparse depth data with or without additional dense RGB data. To study and handle sparsity, we focus on the task of depth completion of sparse depth data but our proposal can also accomplish semantic segmentation with minor adjustments, as we shall demonstrate, and copes with inputs of varying density. Multiple inputs efficiency is proven using sparse depth with dense RGB, which further improves performance.

The proposed method uses a network architecture adapted from NASNet \cite{zoph2017learning} (sec. \ref{sec:net-arch}), with an encoder-decoder for larger receptive field.
We chose not to use a validity mask after an extensive analysis (sec. \ref{sec:validity-mask}) which led to a lighter network with better performance.

A sparse training strategy is proposed to be more robust to varying input density (sec. \ref{sec:sparse-training}). Finally, a late fusion scheme is used when using multiple inputs (sec. \ref{sec:fusion}).\\

\subsection{Network Architecture}
\label{sec:net-arch}

For dense data, state-of-the-art methods in semantic segmentation use encoder-decoder networks or dilated convolutions \cite{chen2018deeplab}. While the latter significantly reduces
the number of parameters, they are ill-conditioned for sparse data as dilated kernel with zeros between weights can miss available pixels. In \cite{uhrig2017sparsity},
a network without any downsampling is used with at most 11x11 kernels, implying a rather small receptive field.
As a consequence, \cite{uhrig2017sparsity} perform similarly as classical local interpolation \cite{ku2018defense} and work less efficiently on very sparse data. As in \cite{ma2018sparse,zhang2018deep} we prefer encoder-decoder with larger receptive field, thus enabling better data completion. 

The encoder part of our network is an adaptation of NASNet \cite{zoph2017learning} which is flexible and very efficient in terms of parameters vs. performance.
We use the mobile version to fit real-time constraints and slightly modify it by removing batch normalization after the first strided convolution layer when the input is sparse.
The latter is necessary because zero values of missing pixels falsify the mean computation of the batchnorm layer.

We build our own custom decoder with transposed convolutions for upsampling, normal convolutions, and copy and concatenate skip connections like in \cite{ronneberger2015u} between the encoder and decoder stages of equivalent resolution.

Numerous choices in this paper were guided by experiments with a small custom encoder-decoder exhibiting acceptable performance on sparse depth but that failed at extracting good features from dense RGB.

\subsection{Analysis of Validity Mask}
\label{sec:validity-mask}

A validity mask \cite{uhrig2017sparsity,luo2016understanding} is a binary matrix of same size as the input data, with ones indicating available input data and zeros elsewhere.
To propagate the mask through the network, \cite{uhrig2017sparsity} use a max pooling of similar kernel size $k$ and stride $s$ than feature convolution ($k, s$). Intuitively, the mask expresses whether the current filter contains at least one valid pixel. The authors further normalize convolution by the number of valid pixels in the filter and rescale valid outputs. 
The drawback of such an approach is that the scaling tends to over-activate highly-upscaled features at lower density.
This problem is similar in spirit to the problem of extension of the valid domain \cite{graham2017submanifold}.
As a consequence, the mask is almost entirely valid at deeper layers.

\begin{figure}
	\centering
	\subfloat[Mask saturation]{\label{fig:validityMaskPlot}\includegraphics[height=.24\linewidth]{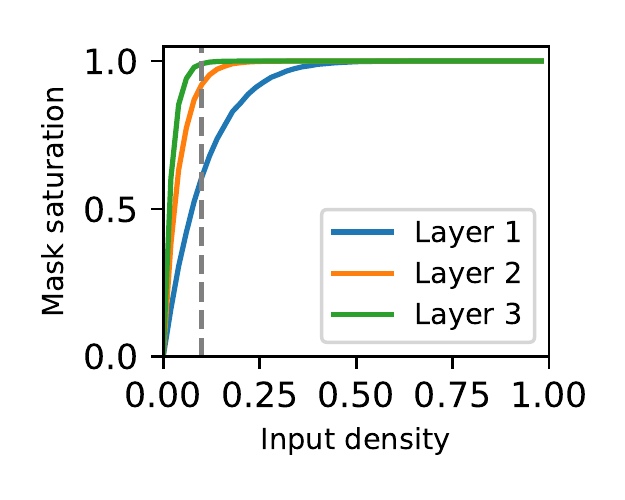}}\hspace{.01\linewidth}
	\subfloat[Mask at 0.1 density]{\label{fig:validityMaskViz}\includegraphics[height=.252\linewidth]{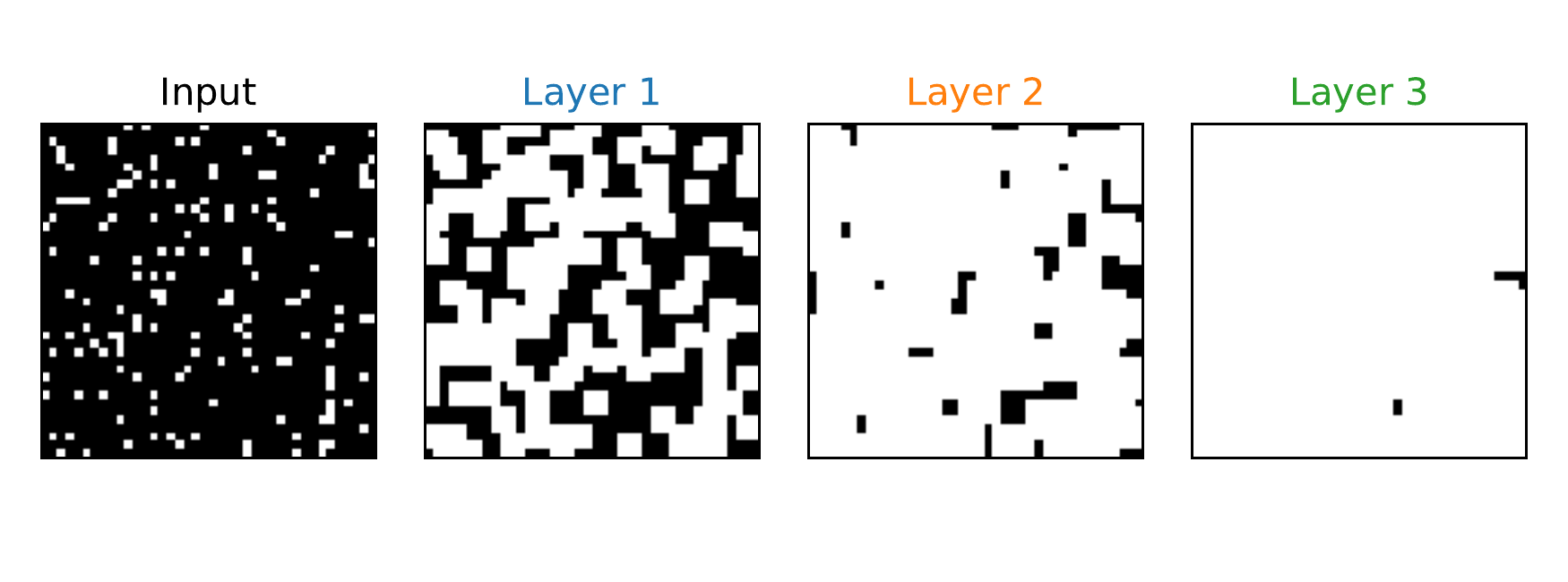}}
	\caption{Saturation of the validity mask for different input densities (\ref{fig:validityMaskPlot}) with max pooling as \cite{uhrig2017sparsity} (here 3 convs: stride 1, kernel 3). For density $\geq{}0.3$, the mask is saturated after first conv. Even at $0.1$ density the spatial location of valid pixels is already mostly lost in deeper layers (\ref{fig:validityMaskViz}).}
	\label{fig:validityMask}
\end{figure}

In fig.~\ref{fig:validityMaskPlot}, we can see
that the mask saturation (the percentage considered as valid) increases with input density as expected, but reaches almost full saturation after only a few layers.
This means that the validity information is quickly lost in the later layers which is visible in fig.~\ref{fig:validityMaskViz} showing an example of such a validity mask and how it is transformed after a few layers.
Another consequence is that the network tends to produce
blurry outputs as seen in fig.~\ref{fig:qualitiativeResultsKittiSynthia}. We interpret that this is a consequence of the normalization phase on the number of valid pixels, which
processes a mask with only one valid pixel in the same way as a fully valid mask. In an attempt to leverage the latter issue, we tested average pooling to preserve the ratio of
valid/invalid pixels in the filter. However, this did not improve results either.

We tested to let the network learn how to use the validity mask by concatenating the actual validity mask channel-wise to the features before each convolution.
While it improves performance on small networks, we observed no improvement with validity masks for large networks such as NASNet.

Tests confirmed our analysis that without any validity mask large networks still manage to learn sparsity invariant features on their own while preserving spatial
information about the validity pixels. 
Consequently, we do not use any validity mask.

\subsection{Sparse Data Training}
\label{sec:sparse-training}

\paragraph{Varying density}

Existing research surprisingly only use fixed density when training although we found that varying synthetic densities within range of $]0, 1]$ naturally helps networks to be invariant to different densities, as discussed in sec. \ref{sec:varyingDensitySynthia}.

Another interesting proposal from \cite{devries2017improved} is to apply rectangles cut-out on data. While this should force the network to use farther away features, it barely improved results in our experiment.

\paragraph{Losses}
For depth completion, we compute the loss only on unobserved pixels available in the ground truth\footnote{In Kitti dataset the ground truth is also sparse as it results of Lidar augmentation using left-right consistency. Read \cite{uhrig2017sparsity} for details.}.
Other strategies such as computing the loss on all pixels (unobserved and observed) or using a weighted sum of observed and unobserved pixels losses, were proven to work less well.
The interest of our choice is to favor learning prediction of the unknown pixels, over learning to reproduce already measured data.

In accordance with \cite{uhrig2017sparsity} we found the $\mathcal{L}_1$ loss to reach slightly better results than $\mathcal{L}_2$ for depth prediction.
We train using inverse depth in [1/km] which corresponds to the inverse mean average error (iMAE) in the Kitti Benchmark.

The final depth $d$ is obtained by reversing the inverse depth $d_{\text{inv}}$ where $d_{\text{inv}} > 0$ and setting the output value to the maximum representable depth $d_{\text{max}}$ 
where the network regresses to $d_{\text{inv}} = 0$ (non-activation). It reads:
\begin{align}\label{eq:depthOutput}
d =
\begin{cases}
	d^{-1}_{\text{inv}}, & \text{for } d_{\text{inv}} > 0\\
	d_{\text{max}}, & \text{for } d_{\text{inv}} = 0
\end{cases}
\end{align}
As optimizer we use Adam with learning rate of 0.001, $\beta_1=0.9$, $\beta_2=0.999$ and $\epsilon=10^{-8}$.

\subsection{Sparse Depth + RGB Fusion}
\label{sec:fusion}
\begin{figure}
	\centering
	\subfloat[Early fusion]{
		\includegraphics[height=0.34\linewidth]{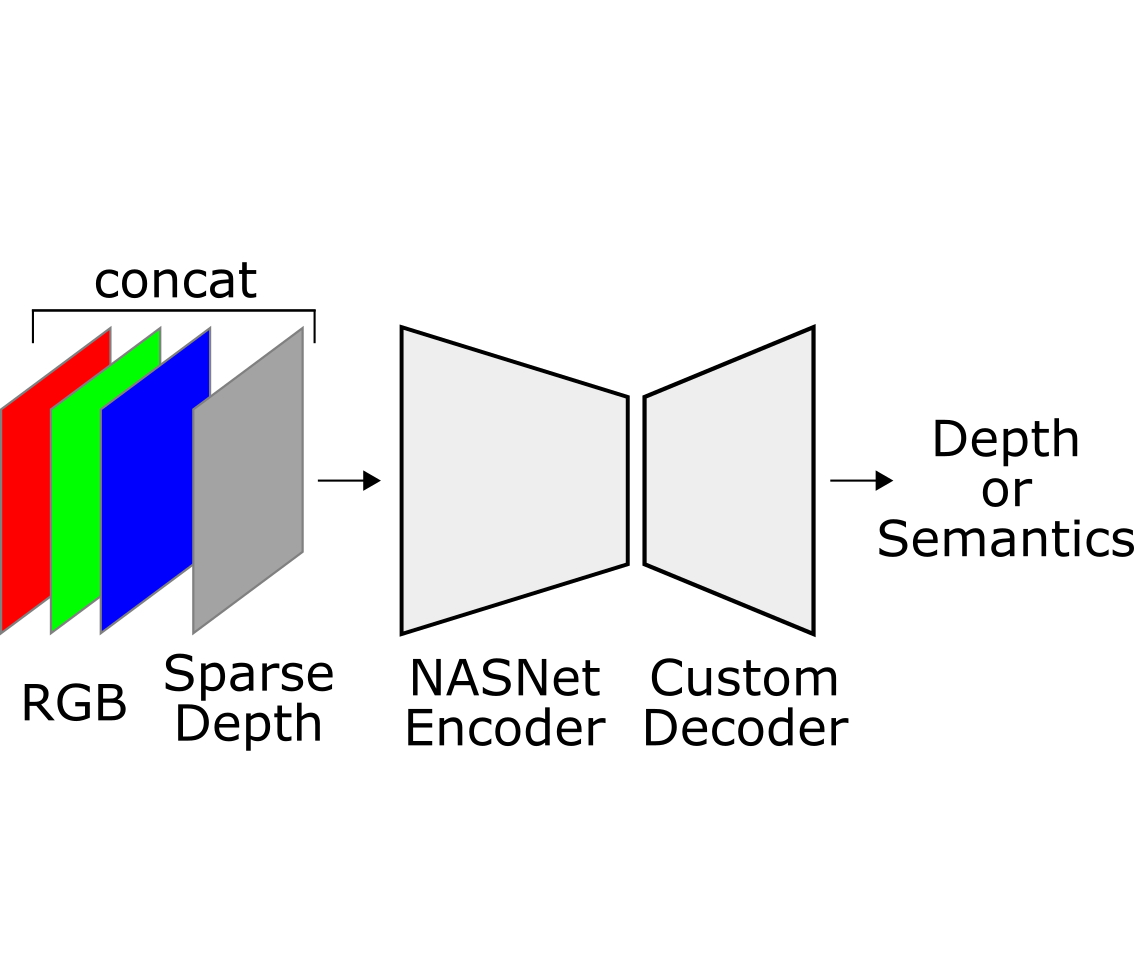}
		\label{fig:earlyFusion}
	}
	\subfloat[Late fusion (ours)]{
		\includegraphics[height=0.34\linewidth]{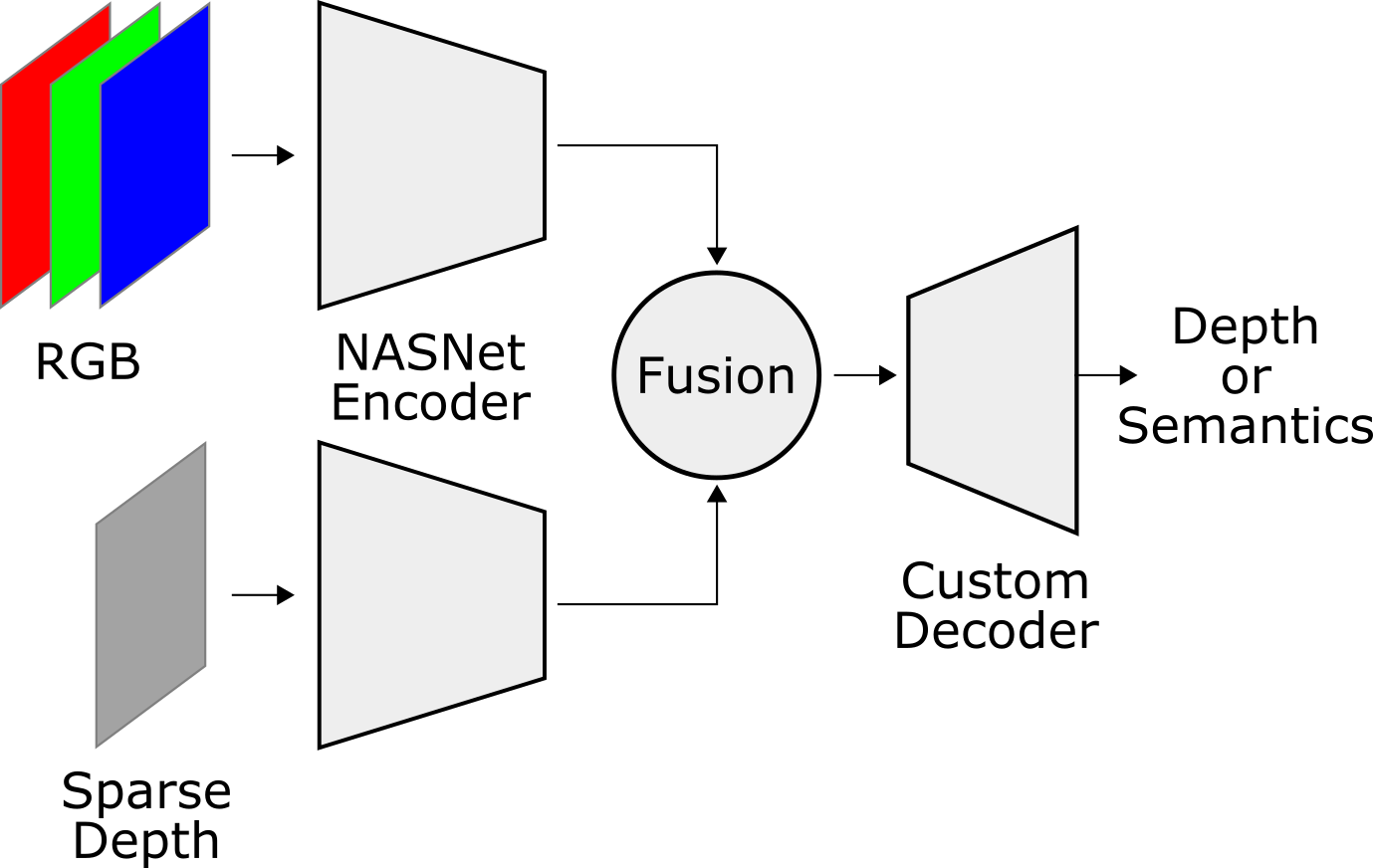}
		\label{fig:lateFusion}
	}
	\caption{Visualization of early fusion (\ref{fig:earlyFusion}) as opposed to our late fusion framework (\ref{fig:lateFusion}).}
\end{figure}

Because the same scene is sensed with a camera and a depth sensor, we want our network to learn to use dense and sparse information jointly for better prediction. A naive strategy consists of averaging separate predictions from each modality. An alternative is to apply an early fusion like in \cite{ma2018sparse}, where modalities are simply concatenated channel-wise and fed to the network (fig.~\ref{fig:earlyFusion}).

However, modalities have different representations (RGB intensities, distance values) and in order to reason from both at the same time, it appears preferable to transform them to a similar feature space before fusing them (known as late fusion fig.~\ref{fig:lateFusion}). A joint representation can be enforced by using element-wise addition of features coming from modality specific encoder networks \cite{chen2017multi}. However, we chose instead to use channel-wise concatenation with a following convolution to allow the two branches to provide information of distinct nature as in \cite{valada2017adapnet}.

In practice, for the task of depth completion using additional dense RGB data requires a particular attention to the choice of architecture and fusion strategy to extract robust enough features from RGB so that fusion actually improves the results. The performance obtained support our choices as for the early vs. late benchmark (sec. \ref{sec:expDepthSynthia} and \ref{sec:semanticSegmentation}).


\section{Experiments}
\label{sec:experiments}

To evaluate our proposal we carried out two experimental tasks: depth completion (sec. \ref{sec:depthCompletion}) and semantic segmentation (sec. \ref{sec:semanticSegmentation}). For both tasks we used either sparse depth (sD) Lidar input, dense RGB input, or a fusion of both and tested on both synthetic and real public datasets described in sec. \ref{sec:datasets}.

For the depth completion task our proposed method can handle input of varying density, reaching better results than others or when trained with a fixed density.
We reached above state-of-the-art on 3 out 4 metrics on the real Kitti Depth Completion Benchmark and our Lidar ablation study shows consistent depth maps reconstructed from only 8 lidar layers.
Finally, for semantic segmentation we prove that our method can handle sparse depth only and significantly improves when fusing RGB and sparse depth compared to the RGB only baseline.

\subsection{Datasets}
\label{sec:datasets}

\paragraph{Synthia}
This dataset built with the Unity game engine provides RGB, depth and semantics for urban and highway scenarios \cite{Ros_2016_CVPR} with pedestrian and cars. 
We use summer sequences 1, 2, 4 and 6 for training and sequence 5 for testing (all views). With our split the training/validation/testing sets contain 28484/1500/6296 frames cropped (bottom/center) and rescaled to 320x160 pixels.

Uniform sparse depth input is simulated by setting different ratios of pixels to zero, which we call pixel density. A density of 0.1 means that 10\% of the pixels are available and 90\% are not.

\paragraph{Kitti}
The Kitti Depth dataset provides RGB, raw lidar data (64 layers HDL-64E) projected into the image plane, and sparse ground truth from the accumulation of lidar point clouds with stereo consistency check \cite{uhrig2017sparsity}.
Comparison against other methods (sec. \ref{sec:depthCompletionReal}) is performed at full-resolution (1216x352), but cropped (bottom/center) and rescaled to 608x160 to reduce training time elsewhere.

Depth being sparse, we use max pooling to downsample it to avoid loosing any points (common resize methods would take zeros into account and corrupt the output).
\paragraph{Cityscapes}
Cityscapes \cite{Cordts2016Cityscapes} provides RGB and stereo disparity from German cities, with 20000 coarse and 3000 fine semantic annotations for training. We resized the data from 2048x1024 to 512x256 for time matters.

\subsection{Depth completion}
\label{sec:depthCompletion}

We first evaluate on synthetic data (Synthia) and then on real data (Kitti). The metrics are from the Kitti Benchmark: Mean Average Error (MAE, $\mathcal{L}_1$ mean over all pixels),  Root Mean Square Error (RMSE, $\mathcal{L}_2$ over all pixels), both in mm, as well as their inverse depth counterparts iMAE and iRMSE in 1/km between prediction $\hat{y}$ and ground truth $y$ averaged over the number of evaluated pixels $N$. 

For the experiment on synthetic data in table \ref{tab:depthCompletionSynthetic} we also report the $\delta$-metric as in \cite{eigen2014depth}, which is the percentage of relative errors inside a certain threshold $\epsilon = \{1.05, 1.10, 1.25, 1.50\}$:
\begin{align}
\delta =  \frac{1}{N}\sum\limits_{i} (\delta_i < \epsilon), \qquad \delta_i = \max\left(\frac{\hat{y_i}}{y_i}, \frac{y_i}{\hat{y_i}}\right)
\end{align}

\subsubsection{Synthetic Data (Synthia)}\label{sec:expDepthSynthia}
\paragraph{Fixed Density}

\begin{table}
	\scriptsize
	\centering
	\begin{tabular}{lrcccr}  
		\toprule
		Input & iMAE & $\delta1.05$ & $\delta1.10$ & $\delta1.25$ & $\delta1.50$ \\
		\midrule
		RGB & 13.56 & 0.56 & 0.69 & 0.85 & 0.92 \\
		sD & 4.05 & 0.86 & 0.91 & 0.95 & 0.97 \\
		RGB~+~sD (Early fusion) & 4.37 & 0.82 & 0.89 & 0.95 & 0.97 \\
		RGB~+~sD (Late fusion) & \textbf{2.96} & \textbf{0.87} & \textbf{0.92} & \textbf{0.96} & \textbf{0.98} \\
		\bottomrule
	\end{tabular}
	\caption{Depth completion on Synthia at input depth density of 0.02. For iMAE lower is better, for $\delta$-metric higher is better, indicated numbers are thresholds. While sparse depth is clearly the most important modality for depth prediction, late fusion can considerably improve the results.}
	\label{tab:depthCompletionSynthetic}
\end{table}

\begin{figure}
	\centering
	\scriptsize
	\setlength{\tabcolsep}{0.05cm}
	\renewcommand{\arraystretch}{0.5}
	\begin{tabular}{cc}
		\toprule
		\textbf{Synthia} & \textbf{Kitti} \\
		\midrule
		\includegraphics[height=0.18\linewidth]{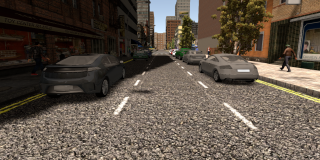} & \includegraphics[height=0.18\linewidth]{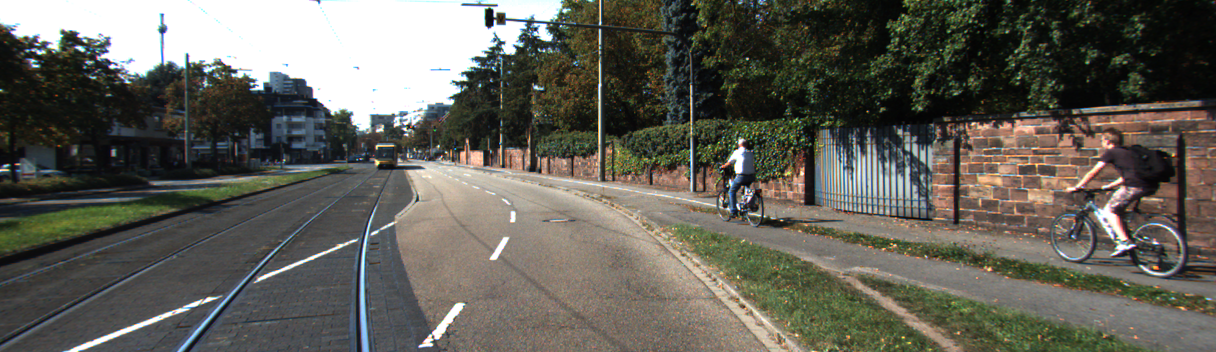} \\
		\multicolumn{2}{c}{RGB input} \\
		\includegraphics[height=0.18\linewidth]{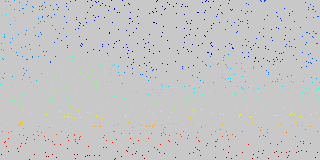} & \includegraphics[height=0.18\linewidth]{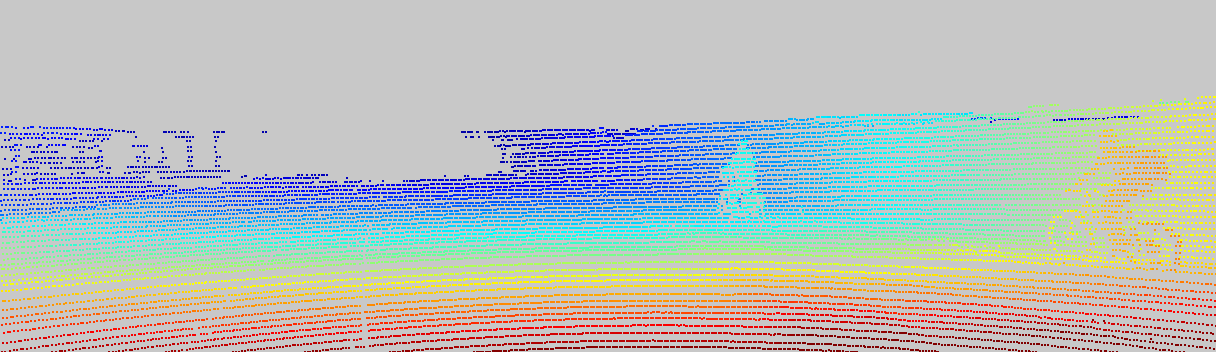} \\
		\multicolumn{2}{c}{sD input} \\
		\midrule
		\includegraphics[height=0.18\linewidth]{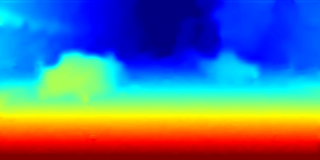} & \includegraphics[height=0.18\linewidth]{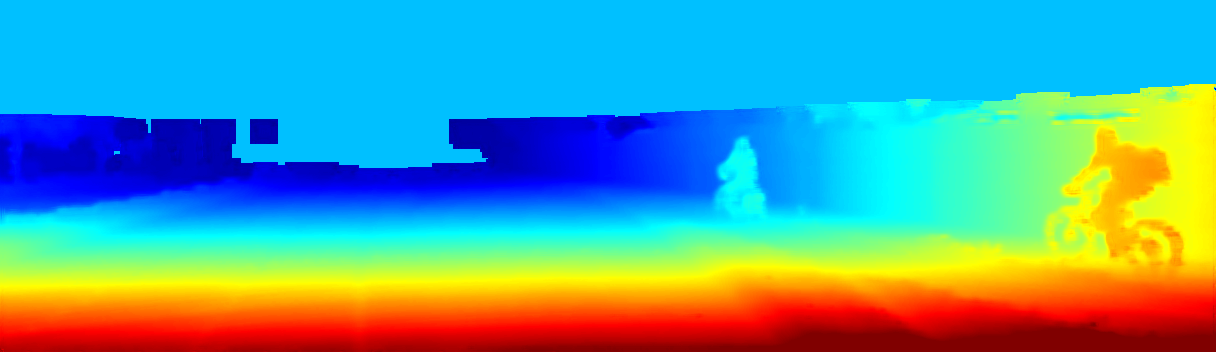} \\
		\multicolumn{2}{c}{Sparse Conv \cite{uhrig2017sparsity}} \\
		\includegraphics[height=0.18\linewidth]{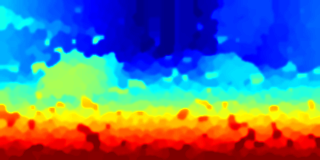} & \includegraphics[height=0.18\linewidth]{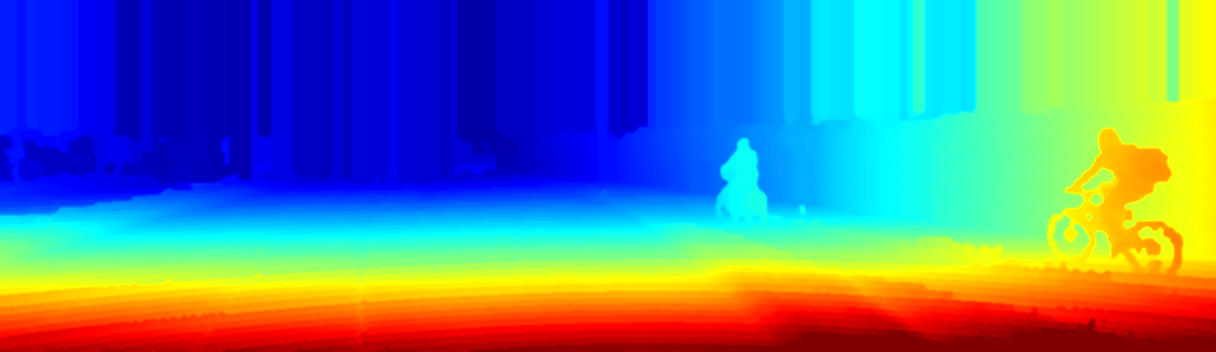} \\
		\multicolumn{2}{c}{IP-Basic \cite{ku2018defense}} \\
		\includegraphics[height=0.18\linewidth]{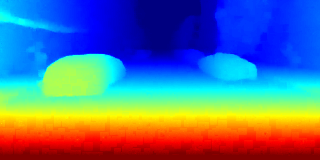} & \includegraphics[height=0.18\linewidth]{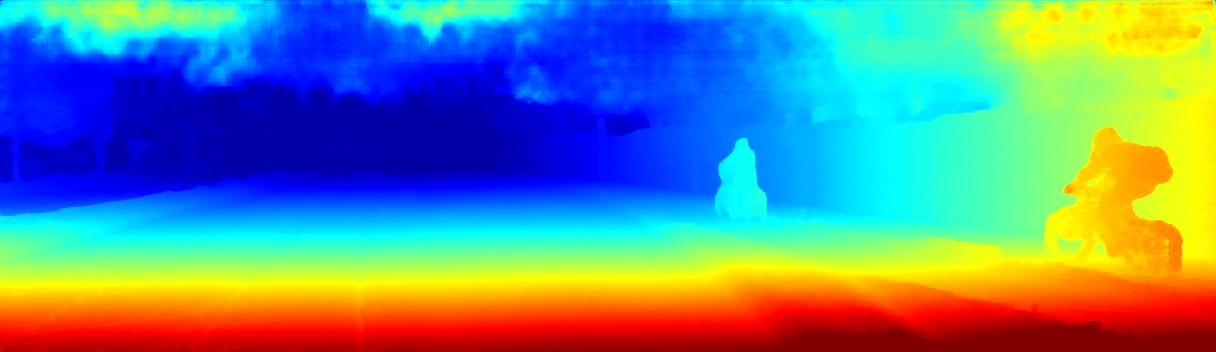} \\
		\multicolumn{2}{c}{Ours sD} \\
		\includegraphics[height=0.18\linewidth]{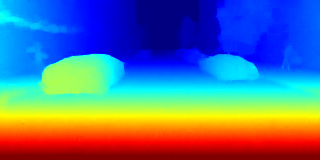} & \includegraphics[height=0.18\linewidth]{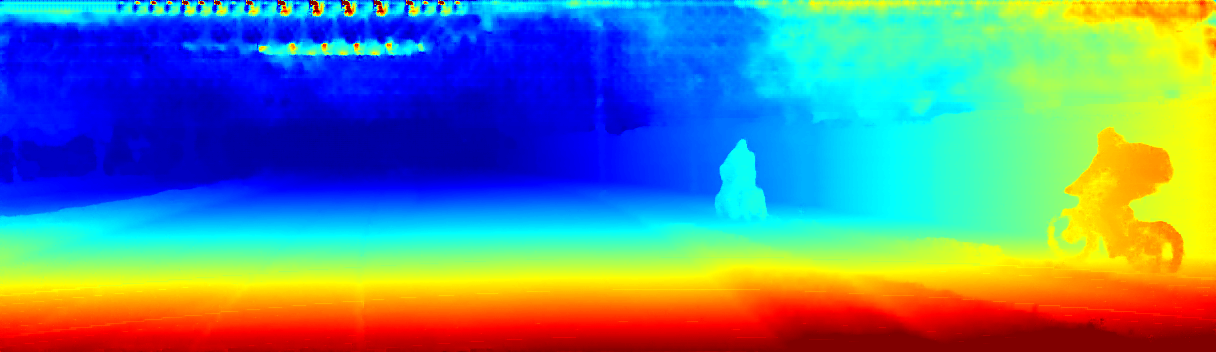} \\
		\multicolumn{2}{c}{Ours RGB~+~sD} \\
		\midrule
		\includegraphics[height=0.18\linewidth]{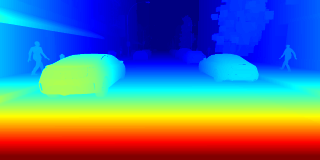} & \includegraphics[height=0.18\linewidth]{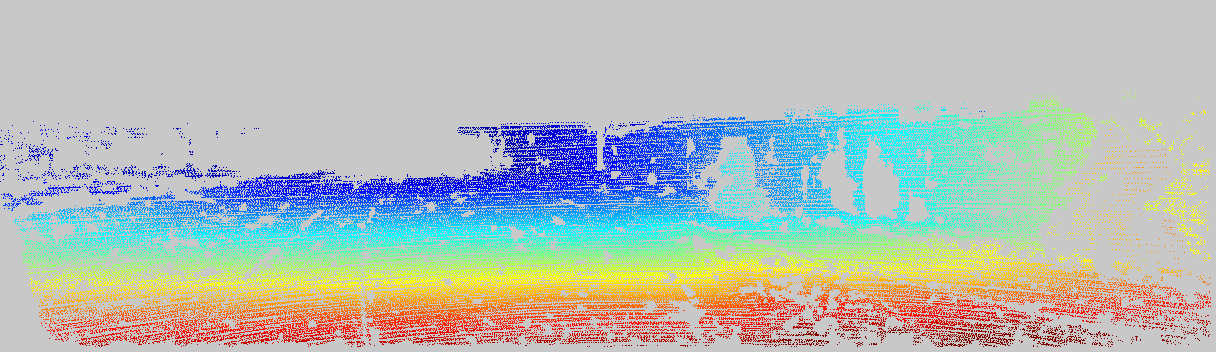} \\
		\multicolumn{2}{c}{Ground truth} \\
	\end{tabular}
	\caption{Qualitative results for depth completion on Synthia (synthetic) at $0.02$ density and Kitti (real) validation set with 64 layers Lidar.}
	\label{fig:qualitiativeResultsKittiSynthia}
\end{figure}

For this experiment we train and evaluate our method on a very low input pixel density of 0.02 (only 2 percent of pixels available), fixed during training but with random valid pixel positions. 
We train for 30 epochs and cherry-pick the best weights and report the iMAE as well as the $\delta$-metric in table \ref{tab:depthCompletionSynthetic}. 
While sparse depth (sD) alone ($4.05$) performs better than RGB alone ($13.56$) - proving thus that our network handles sparse data efficiently - the best results are obtained with the late fusion of RGB~+~sD~($2.96$). 
This is probably because the network can combine the learned visual features and learned geometric features from RGB and very precise sparse depth information. 
Late fusion clearly outperforms early fusion confirming that one network branch per modality is needed to map modalities to a similar feature space before fusion.
Early fusion performs approximately as good as sparse depth only suggesting that the network simply ignores the less informative input modality.
Qualitative results in fig.~\ref{fig:qualitiativeResultsKittiSynthia} exhibit sharp and precise completion for sD and even better for RGB+sD (notice the pedestrians).
While \cite{uhrig2017sparsity} performs decently, \cite{ku2018defense} exhibits sharp but chaotic results due to the very low density.
Note that sparse depth is much more important as modality than RGB, even at this low sparsity level, because the input is effectively a subset of the ground truth.
This effect is even stronger in the case of the Kitti benchmark where the ground truth is constructed with consecutive Lidar measurements. Another consequence is that the network cannot be trained to perform depth completion outside of the field of view of the depth sensor, because ground truth is never available in this area.
This is why in section~\ref{sec:depthCompletionReal} we conduct an ablation study: Lidar layers are removed to test the robustness to lower
density but also the extrapolation capabilities of the inference.

\begin{figure}
	\centering
	\includegraphics[width=0.5\linewidth]{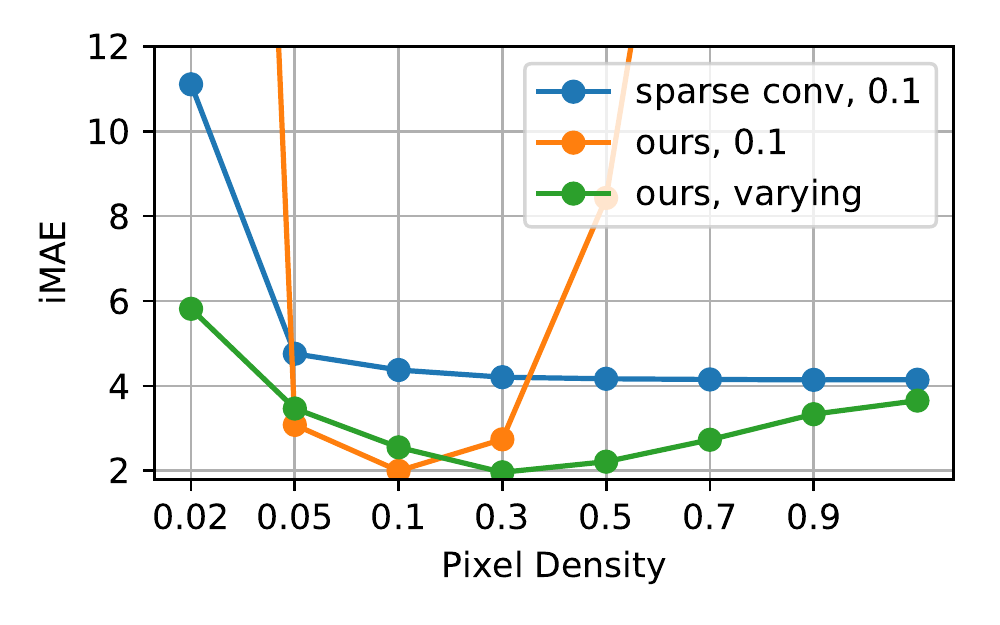}
	\caption{Test results on Synthia for depth completion with sparse depth only input at different density of sparse convolution \cite{uhrig2017sparsity} trained at fixed density of 0.1 (blue), our network trained at fixed density of 0.1 (orange) and our network trained at varying densities between 0 and 1.}
	\label{fig:varyingSparsityPlot}
\end{figure}

\paragraph{Varying Density}\label{sec:varyingDensitySynthia}
We further investigate the influence of different input densities at test time and plot results in fig.~\ref{fig:varyingSparsityPlot} for sparsity invariant CNNs \cite{uhrig2017sparsity} trained as in the paper with a fixed density (here of 0.1), our method trained at fixed density of 0.1, and our method trained at varying density randomly chosen in $]0,1]$ per image. Despite being trained at fixed density, the performance of \cite{uhrig2017sparsity} is almost perfectly stable over different densities at test time, except for the lowest of 0.02.
Our method trained at fixed density of 0.1 achieves much better results for densities $\{0.05, 0.1, 0.3\}$ that are close or equal to the training density. However, the error increases drastically beyond, including the case where more information is given (higher density). The network thus specializes in densities seen during training.
However, when we train our network at varying density between 0 and 1, it gets very robust and we obtain better results than \cite{uhrig2017sparsity} at all densities even at the lowest density of 0.02.

Results demonstrate that our method with varying training density could perform under a large variety of sensor densities which has great implications for Lidar applications.
For the other experiments, where training and test have same densities we use a fixed density to guarantee the best results at test time.

\subsubsection{Real Data (Kitti)}
\label{sec:depthCompletionReal}
\paragraph{Depth Completion Benchmark}
\begin{table}
	\centering
	\scriptsize
	\begin{tabular}{lrcccc}
		Method & Input & iRMSE & iMAE & RMSE & MAE \\
		\toprule
		HMS-Net\_v2* & RGB+sD & 3.90 & 1.90 & \bf{911.49} & 310.14 \\
		HMS-Net* & sD & 3.25 & 1.27 & 976.22 & 283.76 \\
		IP-Basic \cite{ku2018defense} & sD & 3.78 & 1.29 & 1288.46 & 302.60 \\
		SparseConvs \cite{uhrig2017sparsity} & sD & 4.94 & 1.78 & 1601.33 & 481.27 \\
		\midrule
		Ours & sD & 2.60 & 0.98 & 1035.29 & 248.32 \\
		Ours & RGB+sD & \bf{2.17} & \bf{0.95} & 917.64 & \bf{234.81} \\
	\end{tabular}
	\caption{Depth completion performance on the Kitti benchmark. (*~anonymous)}
	\label{tab:depthCompletionKitti}
\end{table}

\begin{figure}
	\scriptsize
	\centering
	\setlength{\tabcolsep}{0.05cm}
	\renewcommand{\arraystretch}{0.5}
	\begin{tabular}{cc}
		\begin{tikzpicture}
		\node (rgb) {\includegraphics[width=0.45\columnwidth]{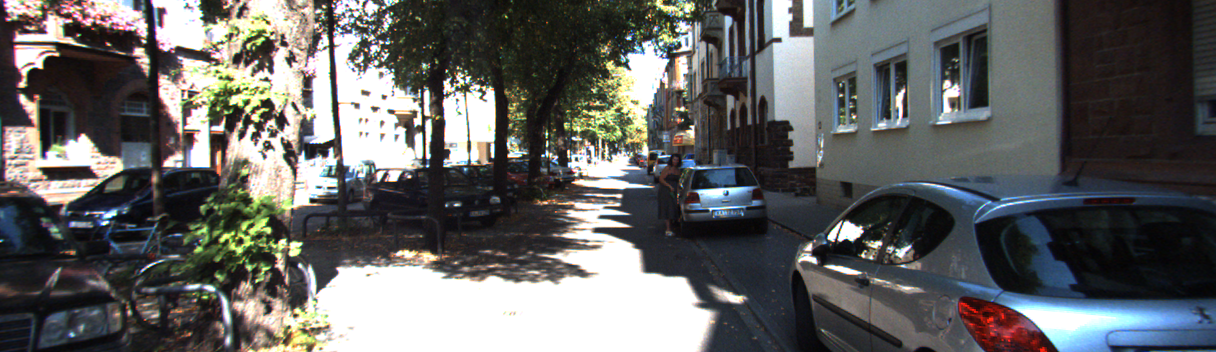}};
		\node[above of=rgb, xshift=1.46cm, yshift=-0.75cm, inner sep=0pt] (rgbzoom) {\includegraphics[width=0.15\columnwidth,trim={80px 0px 936px 180px},clip]{figures/qualitative-results-depth-completion/kitti-benchmark/frames/0000000012_rgb.png}};
		\coordinate (toprect) at (-1.65, 0.0);
		\coordinate (botrect) at (-1.03, -0.540);
		\draw[red](toprect) rectangle (botrect);
		\draw[red] (rgbzoom.north west) rectangle (rgbzoom.south east);
		\end{tikzpicture}
		& \includegraphics[width=0.475\columnwidth]{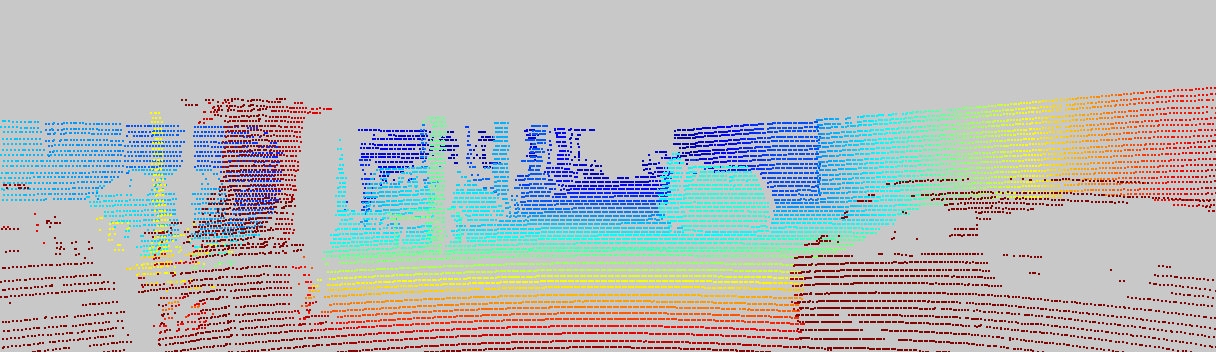} \\
		RGB input & sD input \\
		\begin{tikzpicture}
		\node (rgb) {\includegraphics[width=0.45\columnwidth]{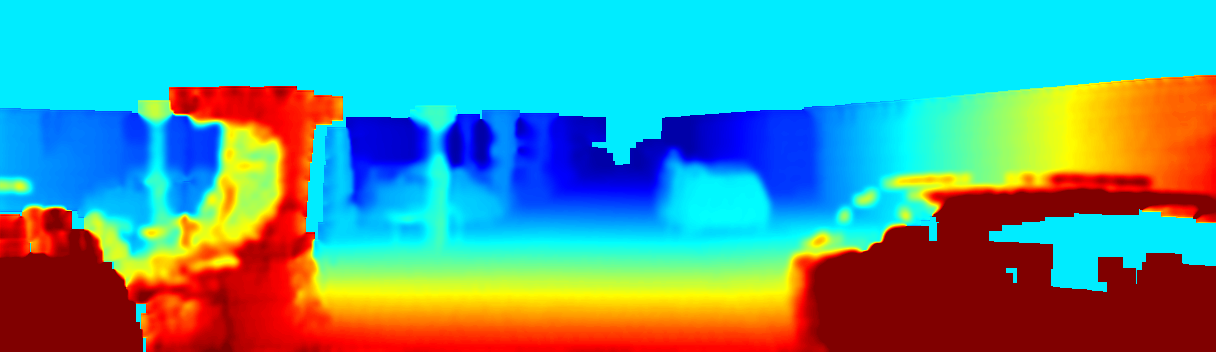}};
		\node[above of=rgb, xshift=1.46cm, yshift=-0.75cm, inner sep=0pt] (rgbzoom) {\includegraphics[width=0.15\columnwidth,trim={80px 0px 936px 180px},clip]{figures/qualitative-results-depth-completion/kitti-benchmark/frames/0000000012_sparcecnn_ourscolorscale-maxdepth_90.png}};
		\coordinate (toprect) at (-1.65, 0.0);
\coordinate (botrect) at (-1.03, -0.540);
\draw[black](toprect) rectangle (botrect);
		\draw[black] (rgbzoom.north west) rectangle (rgbzoom.south east);
		\end{tikzpicture}
		&
		\begin{tikzpicture}
		\node (rgb) {\includegraphics[width=0.45\columnwidth]{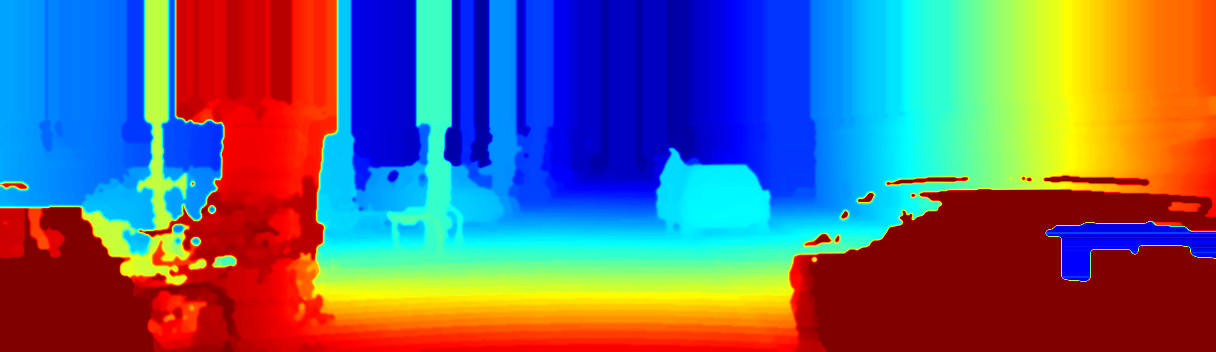}};
		\node[above of=rgb, xshift=1.46cm, yshift=-0.75cm, inner sep=0pt] (rgbzoom) {\includegraphics[width=0.15\columnwidth,trim={80px 0px 936px 180px},clip]{figures/qualitative-results-depth-completion/kitti-benchmark/frames/0000000012_ipbasic_ourscolorscale-maxdepth_90.png}};
		\coordinate (toprect) at (-1.65, 0.0);
\coordinate (botrect) at (-1.03, -0.540);
\draw[black](toprect) rectangle (botrect);
		\draw[black] (rgbzoom.north west) rectangle (rgbzoom.south east);
		\end{tikzpicture}
		\\
		Sparse Conv \cite{uhrig2017sparsity} & IP-Basic \cite{ku2018defense} \\
		\begin{tikzpicture}
		\node (rgb) {\includegraphics[width=0.45\columnwidth]{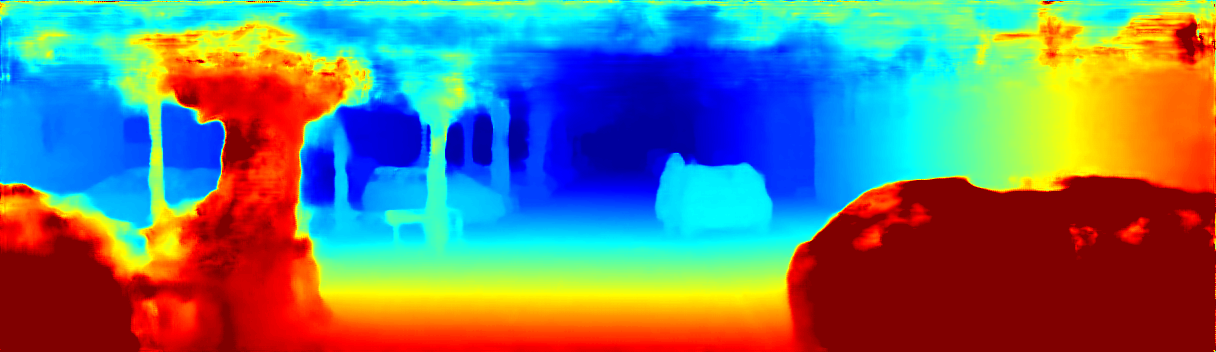}};
		\node[above of=rgb, xshift=1.46cm, yshift=-0.75cm, inner sep=0pt] (rgbzoom) {\includegraphics[width=0.15\columnwidth,trim={80px 0px 936px 180px},clip]{figures/qualitative-results-depth-completion/kitti-benchmark/frames/0000000012_hmsnetv2_ourscolorscale-maxdepth_90.png}};
		\coordinate (toprect) at (-1.65, 0.0);
		\coordinate (botrect) at (-1.03, -0.540);
		\draw[black](toprect) rectangle (botrect);
		\draw[black] (rgbzoom.north west) rectangle (rgbzoom.south east);
		\end{tikzpicture}
		&
		\begin{tikzpicture}
		\node (rgb) {\includegraphics[width=0.45\columnwidth]{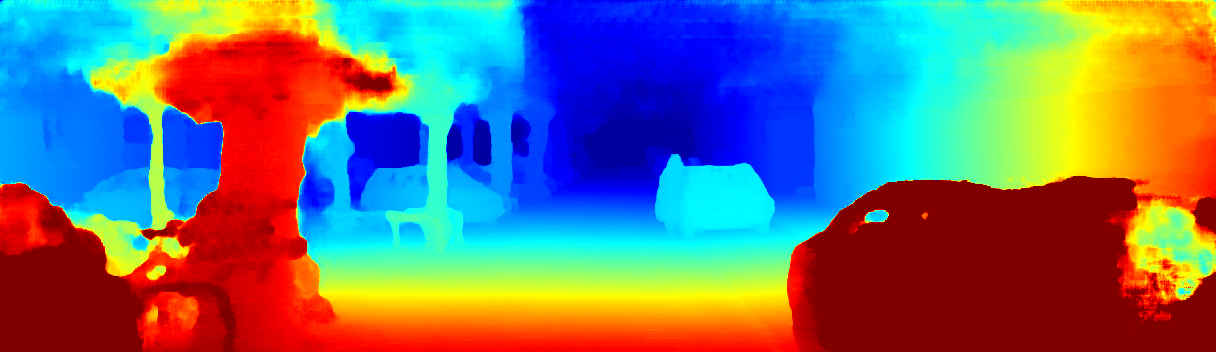}};
		\node[above of=rgb, xshift=1.46cm, yshift=-0.75cm, inner sep=0pt] (rgbzoom) {\includegraphics[width=0.15\columnwidth,trim={80px 0px 936px 180px},clip]{figures/qualitative-results-depth-completion/kitti-benchmark/frames/0000000012_oursrgbsd_ourscolorscale-maxdepth_90.png}};
		\coordinate (toprect) at (-1.65, 0.0);
		\coordinate (botrect) at (-1.03, -0.540);
		\draw[black](toprect) rectangle (botrect);
		\draw[black] (rgbzoom.north west) rectangle (rgbzoom.south east);
		\end{tikzpicture}
		\\
		HMSNetv2* & Ours RGB+sD \\
	\end{tabular}
	\caption{Qualitative results from public Kitti Depth Completion Benchmark (recolored), with an inset zoom on a parked bike. (*~anonymous)}
	\label{fig:qualitiativeResultsKittiBenchmark}
\end{figure}
In table \ref{tab:depthCompletionKitti} we report the best methods from the Kitti benchmark. At the time of submission, we rank first on all published methods. Accounting for anonymous submissions, we rank first on all metrics but the RMSE (third). 
Fig.~\ref{fig:qualitiativeResultsKittiBenchmark} displays the visual output (recolored) from the benchmark website (test set).
Classical morphological interpolation as IP Basic \cite{ku2018defense} is favored by the dense input of the 64 layers lidar. It produces very sharp output but
fails logically to reconstruct shapes when the density is low.
Sparse Convolution \cite{uhrig2017sparsity} does better on shapes by integrating learned priors, but the output is rather blurry and the network is unable to predict
outside of the field of view of the lidar (top part). These are expected counterparts as explained in sec.~\ref{sec:validity-mask}.
HMSNetv2, an anonymous method at the time of submission, reconstructs shapes well, but produces a slightly blurrier output than ours which might help to decrease the RMSE at the expense of the other metrics. The effect is noticeable in the zoom inset. 

Because the ground truth is sparse and only inside the lidar field of view the network is never supervised regarding its prediction at the top of the image. This is better understood looking at results from the validation set (fig.~\ref{fig:qualitiativeResultsKittiSynthia}) as we can display the ground truth along with the results. 
We use full resolution 1216x352, batch size 8 and train 20 epochs.

\paragraph{Lidar Ablation Study}
\label{sec:lidarAblationStudy}
Because the ground truth is obtained from lidar, the depth modality is much more important than RGB. To compensate for that and to give a clue which depth map precision can be obtained with fewer layer lidars, the input can be reduced from the original: we subsampled the 64 layers Velodyne data to simulate 8, 16 and 32 layers\footnote{We subsampled every 2nd, 4th, and 8th layer to simulate 32, 16 and 8 layers, untwisted data linearly using the car speed from IMU, and projected into the camera image plane.}.

\begin{figure}
	\centering
	\scriptsize
	\subfloat[iMAE error]{\includegraphics[width=0.5\linewidth]{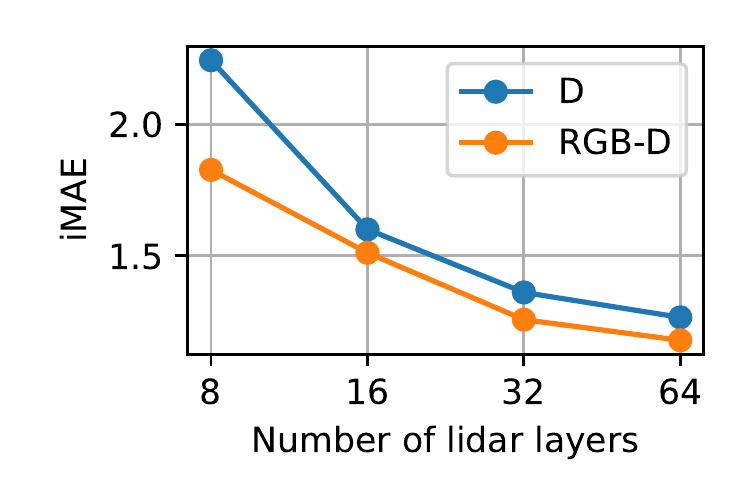}\label{fig:lidarAblationPlot}}

	\subfloat[Qualitative results]{
	\setlength{\tabcolsep}{0.05cm}
        \renewcommand{\arraystretch}{0.5}
	\begin{tabular}{cc}
	 \toprule
	 \textbf{8 layers (6 visible)} & \textbf{32 layers (23 visible)} \\
	 \midrule
	 \includegraphics[width=0.48\columnwidth]{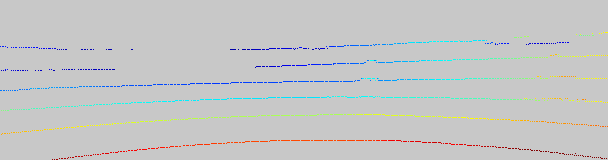} & \includegraphics[width=0.48\columnwidth]{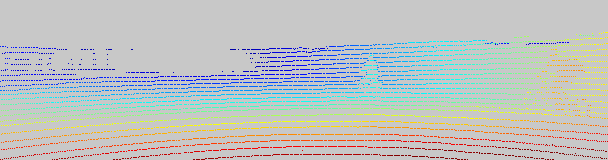} \\
	 \multicolumn{2}{c}{sD input} \\
	 \midrule
	 \includegraphics[width=0.48\columnwidth]{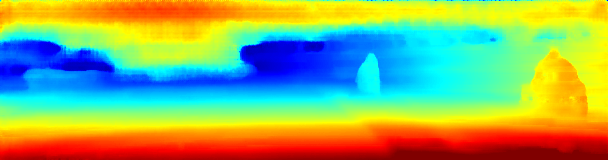} & \includegraphics[width=0.48\columnwidth]{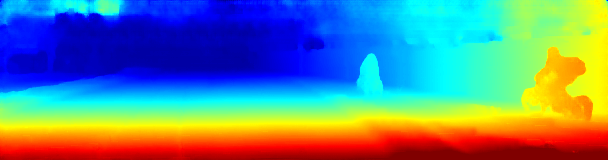} \\
	 \multicolumn{2}{c}{Ours sD} \\
	 \includegraphics[width=0.48\columnwidth]{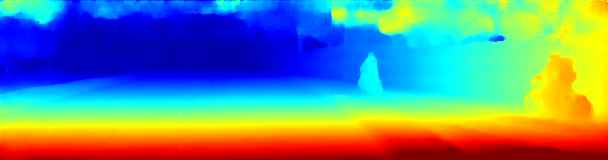} & \includegraphics[width=0.48\columnwidth]{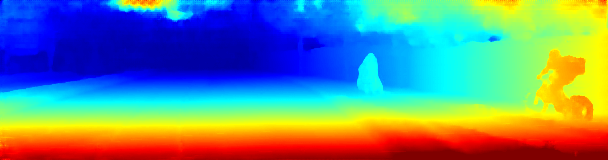} \\
	 \multicolumn{2}{c}{Ours RGB+sD} \\
	\end{tabular}
	}

	\caption{Depth completion with simulated fewer layer lidars (downsampling of 64 layers input). Even with only 8 layers our method completes the depth map remarkably.}
	\label{fig:lidarAblation}
\end{figure}

In fig.~\ref{fig:lidarAblationPlot} we can see that by decreasing the number of input layers the dense depth map prediction deteriorates as expected. Interestingly, the RGB input always improves especially at lower densities.
Qualitative results in fig.~\ref{fig:lidarAblation} show remarkable output even with only 8 layers - i.e. $0.008$ density - (note the bike in the foreground).

\subsection{Semantic Segmentation}
\label{sec:semanticSegmentation}
In this section we investigate how sparse depth can improve semantic segmentation by evaluating our method on synthetic data (Synthia, 13 classes) and real data (Cityscapes, 19 classes). We do not use the Kitti segmentation benchmark, because the corresponding lidar point clouds are not provided and the number of training images is very small (200).

Note that our goal is not to improve state-of-the-art results, but rather to compare against a RGB-only baseline. In the literature semantic segmentation is usually carried out with RGB only and we prove here our method outperforms the baseline using additional sparse depth data.

First, to adapt our network to semantic segmentation we modify the last layer to output the probabilities for all classes (Softmax) instead of the 1-channel depth regression, and train with a cross entropy loss. 
We report the mean Intersection over Union (IoU), first computed per class and then averaged over classes.

\begin{table}
	\centering
	\scriptsize
	\subfloat[Synthia (synthetic)]{
		\begin{tabular}{lr}  
			\toprule
			Input & IoU \\
			\midrule
			RGB (baseline) & 63.47 \\
			sD & 57.10 \\
			RGB~+~sD (Early Fusion) & 65.68 \\
			RGB~+~sD (Late Fusion) & \textbf{70.74} \\
			\bottomrule
		\end{tabular}
		\label{tab:semanticSegmentationSynthia}
	}
	\subfloat[Cityscapes (real)]{
		\begin{tabular}{lr}  
			\toprule
			Input & IoU \\
			\midrule
			RGB (baseline) & 50.13 \\
			sD & 44.18 \\
			RGB~+~sD (Early Fusion) & 50.10 \\
			RGB~+~sD (Late Fusion) & \textbf{57.82} \\
			\bottomrule
		\end{tabular}
		\label{tab:semanticSegmentationCityscapes}
	}
	\caption{Semantic segmentation on Synthia (\ref{tab:semanticSegmentationSynthia}, 0.3 uniform sparse depth) and Cityscapes (\ref{tab:semanticSegmentationCityscapes}, stereo depth) exhibit similar performance. RGB being the main modality but fusion with sparse depth always improves performance.}
\end{table}

\begin{figure}
	\centering
	\scriptsize
	\setlength{\tabcolsep}{0.05cm}
	\renewcommand{\arraystretch}{0.5}
	\begin{tabular}{cc}
		\toprule
		\textbf{Synthia} & \textbf{Cityscapes} \\
		\midrule
		\includegraphics[width=0.48\columnwidth]{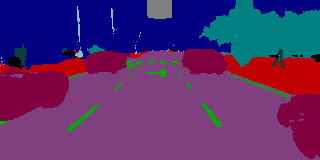} & \includegraphics[width=0.48\columnwidth]{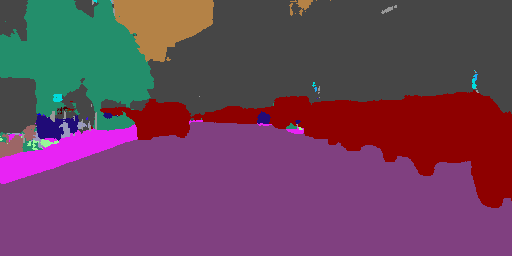} \\
		\multicolumn{2}{c}{RGB baseline} \\
		\includegraphics[width=0.48\columnwidth]{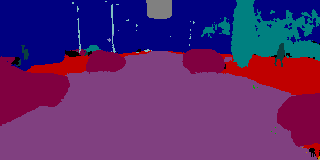} & \includegraphics[width=0.48\columnwidth]{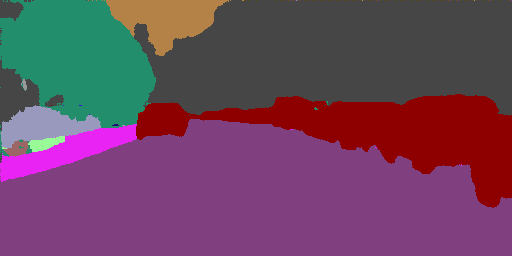} \\
		\multicolumn{2}{c}{Ours sD} \\
		\includegraphics[width=0.48\columnwidth]{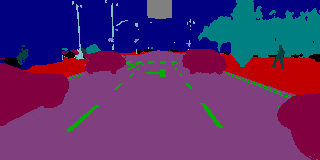} & \includegraphics[width=0.48\columnwidth]{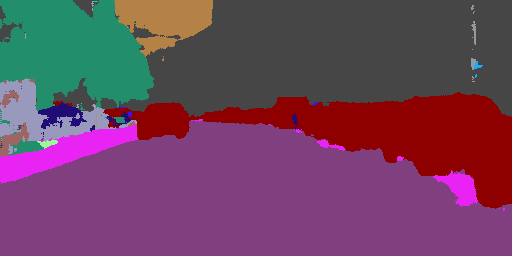} \\
		\multicolumn{2}{c}{Ours RGB+sD} \\
		\midrule
		\includegraphics[width=0.48\columnwidth]{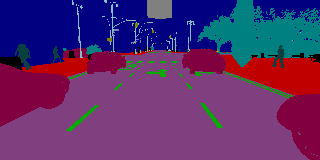} & \includegraphics[width=0.48\columnwidth]{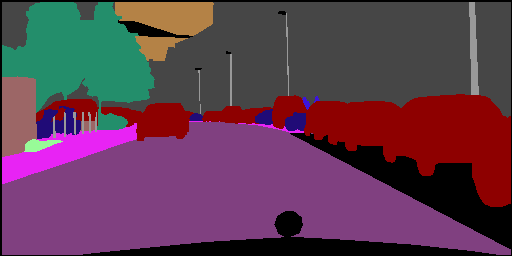} \\
		\multicolumn{2}{c}{Ground truth} \\
	\end{tabular}
	\caption{Qualitative results  for semantic segmentation on Synthia (synthetic, 0.3 sD density) and Cityscapes (real). Note markings that cannot be predicted when using only sD input.}
	\label{fig:qualitiativeResultsSemanticSegmentation}
\end{figure}

\subsubsection{Synthetic Data (Synthia)}\label{sec:semanticSegmentationSynthetic}
For sparse depth, we use a pixel density of 0.3 close to the 64 Velodyne lidar (0.27 inside the FOV) and trained for 30 epochs with a batch size of 16.
The results in table \ref{tab:semanticSegmentationSynthia} indicate naturally that texture and intensity from RGB data carry more semantic information than depth. 
As for depth completion, our RGB+sD late fusion is proved best while early fusion fails to integrate depth features and reaches only slightly better output as RGB only. 
In fig.~\ref{fig:qualitiativeResultsSemanticSegmentation}, sparse depth only shows very acceptable results but fails as expected on lane markings and far away buildings. 
With our late RGB+sD fusion the network better reconstructs the shape of cars.

\subsubsection{Real Data (Cityscapes)}
On real data (see table \ref{tab:semanticSegmentationCityscapes}), we obtain similar results relatively to Synthia although the depth data is different in Cityscapes: almost dense, unscaled disparity from stereo camera instead of metric distance.
This proves the robustness of our proposed method.
We trained 50 epochs on coarse labels and then 50 epochs on fine ground truth using a batch size of 16.
Qualitative results in fig.~\ref{fig:qualitiativeResultsSemanticSegmentation} are similar to synthetic data with satisfying result although thin structures like light poles are never segmented.
The network uses together dense RGB and sparse depth for better segmentation.
\\

Our findings demonstrate that sparse depth can directly be input into a network for semantic segmentation and possibly other tasks such as object detection without first generating the dense depth map like is commonly done in RGB-D networks as for example in \cite{gupta2014learning}.
Additionally, our method works with various densities and sparsity types (Fig.~\ref{fig:sparseDepthExamples}) given that the latter does not change between train and test.

\section{Conclusion}

We presented a method to handle sparse depth data, using a modified NASNet \cite{zoph2017learning} with our decoder, a sparse training strategy and a late fusion scheme for dense RGB + sparse depth.
Following our study of sparse data and validity mask, we do not use any additional mask proving that the network learns sparsity invariant features by itself.

Our results on depth completion outperform all published methods on the Kitti benchmark and are qualitatively remarkable with only 8 layers lidar.
Changing only the last layer, we also performed semantic segmentation on synthetic and real datasets.

The proposed method is proven to efficiently fuse dense RGB and sparse depth. This should benefit all vision tasks using inputs with various densities. Our future work will target application to 3D object detection.

{\small
\bibliographystyle{ieee}
\bibliography{egbib}
}

\end{document}